\documentclass[twoside,11pt]{article}

\usepackage{blindtext}

\usepackage[preprint]{jmlr2e}

\usepackage[autonum, blackhypersetup]{shortex}
\usepackage[english]{babel}
\usepackage{bbm}
\usepackage{caption}
\usepackage{cases}
\usepackage{enumerate}
\usepackage{epsf}
\usepackage{float} 
\floatstyle{plaintop}
\restylefloat{table}
\usepackage{forest}
\usepackage{hyperref}
\usepackage{mathtools}
\usepackage{makecell}
\usepackage{multirow}
\usepackage{pdfpages}
\usepackage{subcaption}
\usepackage{xcolor}
\usepackage{algorithm}
\usepackage{algpseudocode}

\DeclareMathOperator*{\argmin}{arg\,min}
\DeclareMathOperator{\Exp}{E}
\DeclareMathOperator{\Var}{Var}

\DeclarePairedDelimiter{\floor}{\lfloor}{\rfloor}

\newcommand{\reals}{\mathbb{R}}

\newcommand{\code}[1]{\code{#1}}
\def\*#1{\mathbf{#1}}

\definecolor{juliablue}{RGB}{0, 154, 250}
\definecolor{juliaorange}{RGB}{255, 165, 0}
\definecolor{juliared}{RGB}{227, 111, 71}

\usepackage{lastpage}
\jmlrheading{}{}{1-\pageref{LastPage}}{}{}{}{Nikola Surjanovic, Andrew Henrey, and Thomas M.~Loughin}

\ShortHeadings{Alpha-Trimming: Locally Adaptive Tree Pruning for Random Forests}{Surjanovic, Henrey, and Loughin}
\firstpageno{1}

\begin{document}

\title{Alpha-Trimming: Locally Adaptive Tree Pruning \\ for Random Forests}

\author{\name Nikola Surjanovic \email nikola.surjanovic@stat.ubc.ca \\
       \addr Department of Statistics\\
       University of British Columbia\\
       Vancouver, BC V6T 1Z4, CAN
       \AND 
       \name Andrew Henrey \email andrew.henrey@gmail.com \\
       \addr Vancouver, BC, CAN
       \AND
       \name Thomas M. Loughin \email tloughin@sfu.ca \\
       \addr Department of Statistics and Actuarial Science\\
       Simon Fraser University\\
       Burnaby, BC V5A 1S6, CAN}
\editor{}

\maketitle

\begin{abstract}
We demonstrate that adaptively controlling the size of individual regression trees in a random forest can 
improve predictive performance, contrary to the conventional wisdom that trees should be 
fully grown. A fast pruning algorithm, \emph{alpha-trimming},
is proposed as an effective approach to pruning trees within a random forest, where 
more aggressive pruning is performed in regions with a low signal-to-noise ratio.
The amount of overall pruning is controlled by adjusting the weight on an information 
criterion penalty as a tuning parameter, with the standard random forest being a special case of 
our alpha-trimmed random forest. 
A remarkable feature of alpha-trimming is that its tuning parameter 
can be adjusted without refitting the trees in the random forest once the trees 
have been fully grown once.
In a benchmark suite of 46 example data sets, mean squared prediction error
is often substantially lowered by using our pruning algorithm and is never 
substantially increased compared to a random forest with fully-grown trees at default  
parameter settings.
\end{abstract}



\section{Introduction}
\label{sec:intro}
Regression trees have found utility as base learners in ensembles owing to their 
flexibility, near-unbiasedness, and the speed of the most common algorithm for their 
fitting \citep{hastie2017elements}. One ensemble of regression trees, the random forest (RF), computes 
average predictions from a large number of fully-grown trees fitted to bootstrapped versions 
of the data \citep{breiman2001RFs}. The bootstrapping procedure injects mild randomness into the original data, allowing the averaging process to 
reduce the variability of the predictions \citep{hastie2017elements}. 
However, predictions among trees still tend to be highly correlated due to their dependence 
on common features, thus limiting the effectiveness of the averaging.  
The correlation among predictions is often reduced by using only a random 
subset of features as candidates for 
each successive split in each tree during construction. 

Random forests have proven to be worthy prediction machines in a wide range 
of applications \citep{hastie2017elements,polishchuk2009application,belgiu2016random}. 
While the averaging process of a RF can reduce 
prediction variance compared to individual trees 
in the ensemble, it does not reduce any bias. 
Therefore, it is generally recommended 
that individual trees be grown very large, with few observations in each terminal node.  This maximally reduces the potential bias of the predictions at the cost of increasing their 
variance. Indeed, while the tree size is a potential tuning parameter within a RF, few 
sources recommend tuning tree sizes but instead emphasize tuning the fraction of 
variables that are to be considered at each split.
For instance, \cite{hastie2017elements} remark that 
``using full-grown trees seldom costs much, and results in
one less tuning parameter.'' Further, they mention that 
\cite{segal2004machine} ``demonstrates small gains in performance by
controlling the depths of the individual trees grown in random forests.''

\begin{figure}[t]
    \centering
    \begin{subfigure}{.488\textwidth}
      \centering
      \includegraphics[width=\linewidth]{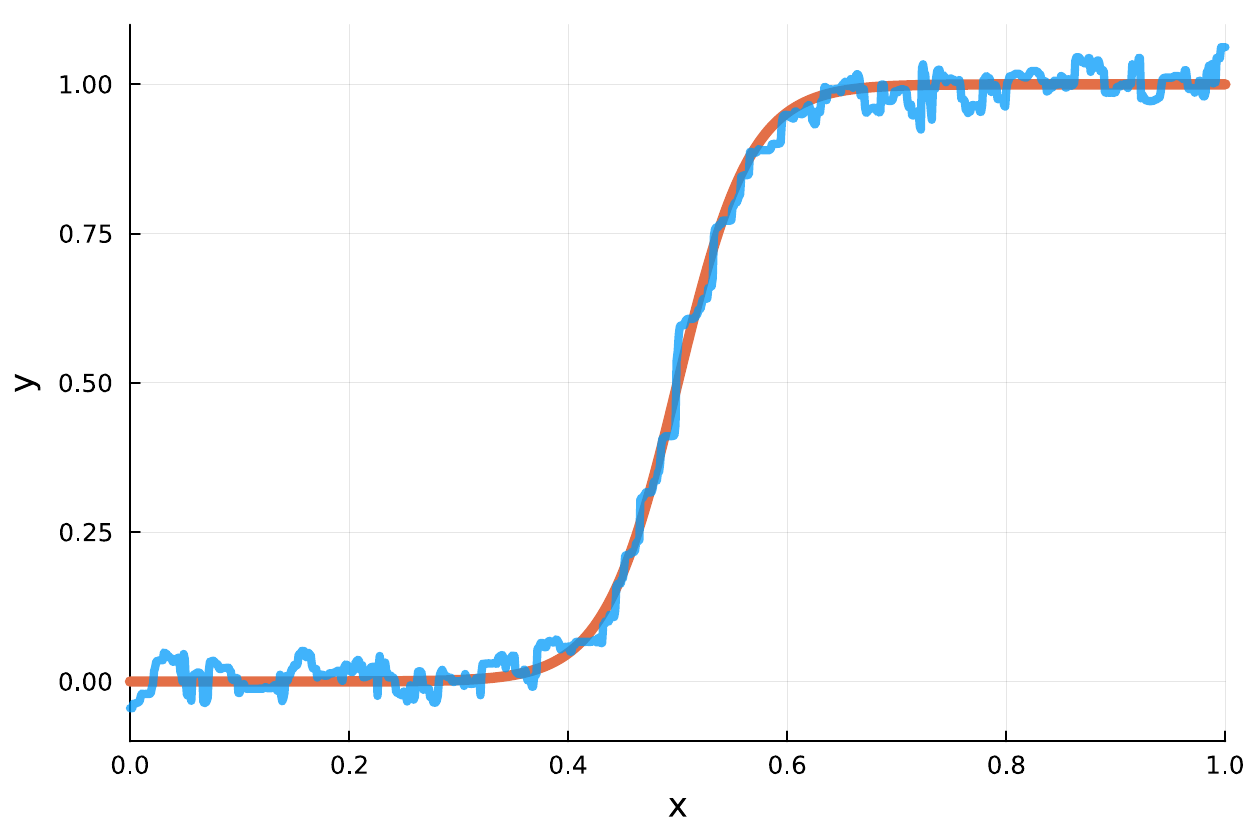}
    \end{subfigure}
    \begin{subfigure}{.488\textwidth}
      \centering
      \includegraphics[width=\linewidth]{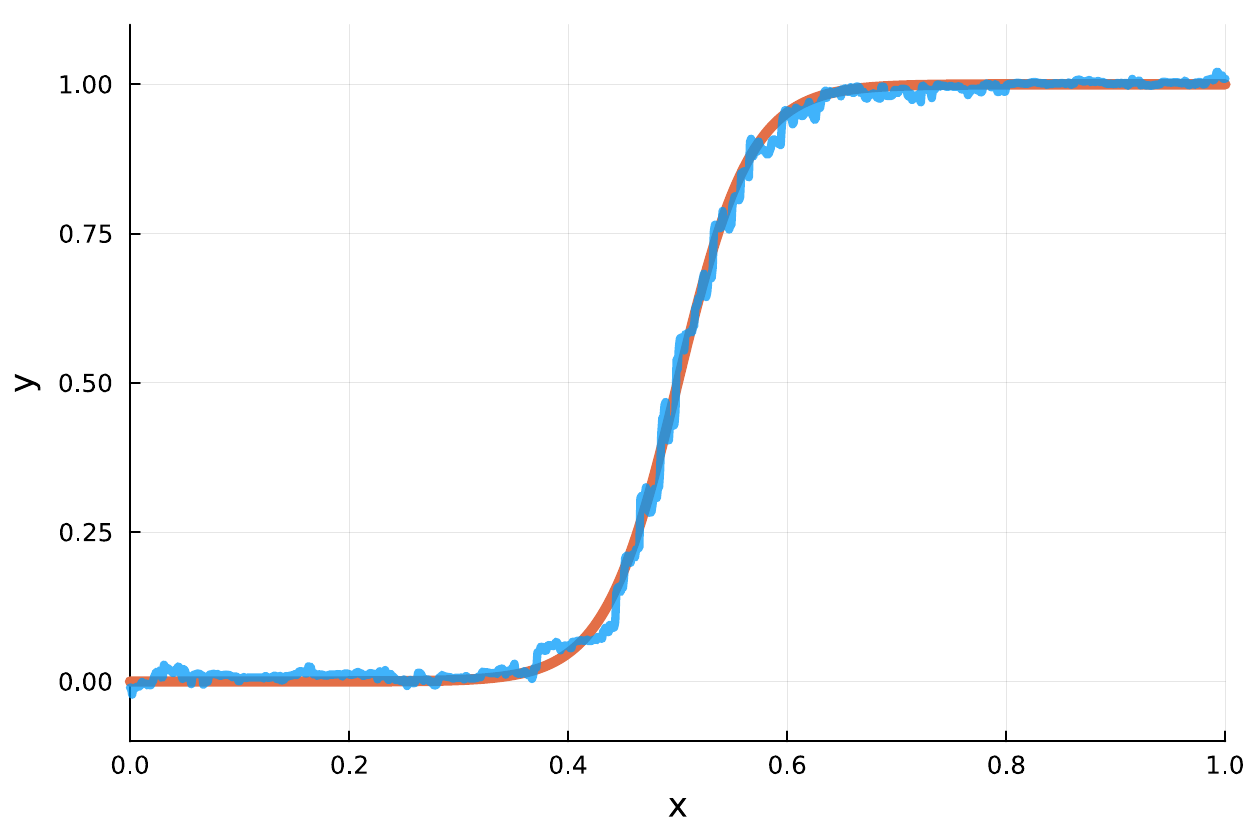}
    \end{subfigure}
    \caption{A logistic regression curve (red) with estimated regression curves (blue). 
    \textbf{Left:} predictions from a random forest with tuned global node size. 
    \textbf{Right:} predictions from our proposed alpha-trimmed random forest with local 
    node size selection.
    The alpha-trimmed random forest approximates the true regression curve well
    in regions with both a high and low signal-to-noise ratio.}
    \label{fig:logistic_example}
\end{figure}

In contrast to this conventional wisdom,
work by \cite{henrey2016statistical} and \cite{zhou2019pruning} finds
that growing large trees leads to predictions that are overly variable in regions where 
the response surface is flat or where there is a low signal-to-noise ratio (SNR). In such 
cases, the bias-variance trade-off favours building trees with larger terminal nodes in 
these regions. 
Thus, it is apparent that some consideration should be given to terminal
node sizes in random forests as a function of the SNR in the local region.  

To vary the size of trees in a RF, a very blunt instrument is to fix
either the tree depth or the minimum number of observations required in order to consider
a further split. However, depending on the tree-building algorithm in the RF 
implementation, these approaches may simply make trees globally smaller without taking 
into account the shape of the response surface in different regions of the feature 
space. To demonstrate this deficiency, we simulated 1,000 
observations from a population with a logistic mean curve and standard normal 
errors. \cref{fig:logistic_example} (a) shows the true mean curve and the 
estimated regression surface using a random forest with tuned node 
size.
While predictive performance appears adequate in some regions of the variable space, 
predictions on the flat regions of the regression curve appear to have rather 
high variability, which is undesirable.

For regression functions where the SNR is variable, a method is needed that can \textit{adaptively}
adjust the size of the trees to the shape of the response surface, allowing more splits to take 
place in regions where the SNR is high and fewer splits in regions where the SNR is low. 
The usual tree-pruning algorithm of \cite{breiman1984classification} should be 
able to do this in theory, but 
applying cross-validation (CV) to individual trees in a RF can be time-consuming.  
Furthermore, within-tree CV considers a prediction performance criterion that balances 
bias and variance for each tree separately, rather than for the entire ensemble. 
For instance, the RF might perform better when predictions from individual trees are more 
variable than would be optimal for prediction using a single tree. 

With these points in mind, we develop a novel 
tree-pruning algorithm that optimizes the predictive performance of the whole forest, yet 
is sensitive to the local node-size requirements of the response surface. Potential 
pruning locations within each tree are determined using an algorithm based on an
information criterion derived for this purpose.  
The amount of pruning applied is controlled by a tuning parameter, $\alpha$, that can be 
easily selected using the RF's out-of-bag error. Because the amount of pruning needed is 
often less than what would normally be applied to an individual regression tree, we refer 
to the algorithm as \emph{alpha-trimming} to indicate that we generally only \textit{trim}
the trees slightly rather than \textit{prune} large portions. 
\cref{fig:logistic_example} (b) shows the performance of the 
alpha-trimmed RF on the simulated data from the logistic curve. Note that we are able to estimate both
low- and high-SNR regions of the regression function well and with low variance.

In \cref{sec:preliminaries} we introduce some notation required to formally study 
our pruning algorithm for trees in a random forest. We develop our 
information-based pruning algorithm for a single tree and then describe how it is extended 
to RFs in \cref{sec:alpha-trimming}. We also introduce the tuning parameter 
$\alpha$ into the algorithm to adaptively control the amount of trimming that is applied 
to the ensemble, and show how to select it without ever refitting any of the 
regression trees. 
\cref{sec:alpha-trimming} additionally contains theoretical results concerning
consistency for our modified Bayesian information criterion when applied to 
simple tree models.
A simulation study is described in \cref{sec:sim_design} and the 
results are presented in \cref{sec:sim_results}. We demonstrate the predictive 
performance of the new alpha-trimmed RF algorithm relative to RFs with fully-grown trees. 
We end with a discussion in \cref{sec:conclusion}. The proofs of theoretical results 
and additional experimental details can be found in the appendices.

\section{Preliminaries and notation}
\label{sec:preliminaries}
Let $(X, Y)$ denote a 
pair consisting of features $X$ and a response variable $Y$.
We assume that $X \in \reals^d$ and $Y \in \reals$. We denote an i.i.d.~random
sample of such pairs of random variables by $(X_i, Y_i)$ for $i=1,\ldots,n$, where $n$ 
is the sample size. Realizations of these random variables are written in lowercase as 
$(x_i, y_i)$. Throughout, we assume that $Y$ is normally distributed
conditional on a given value of $X = x$.

\begin{figure}[t]
  \centering
  \begin{subfigure}{.32\textwidth}
    \centering
    \includegraphics[width=\linewidth]{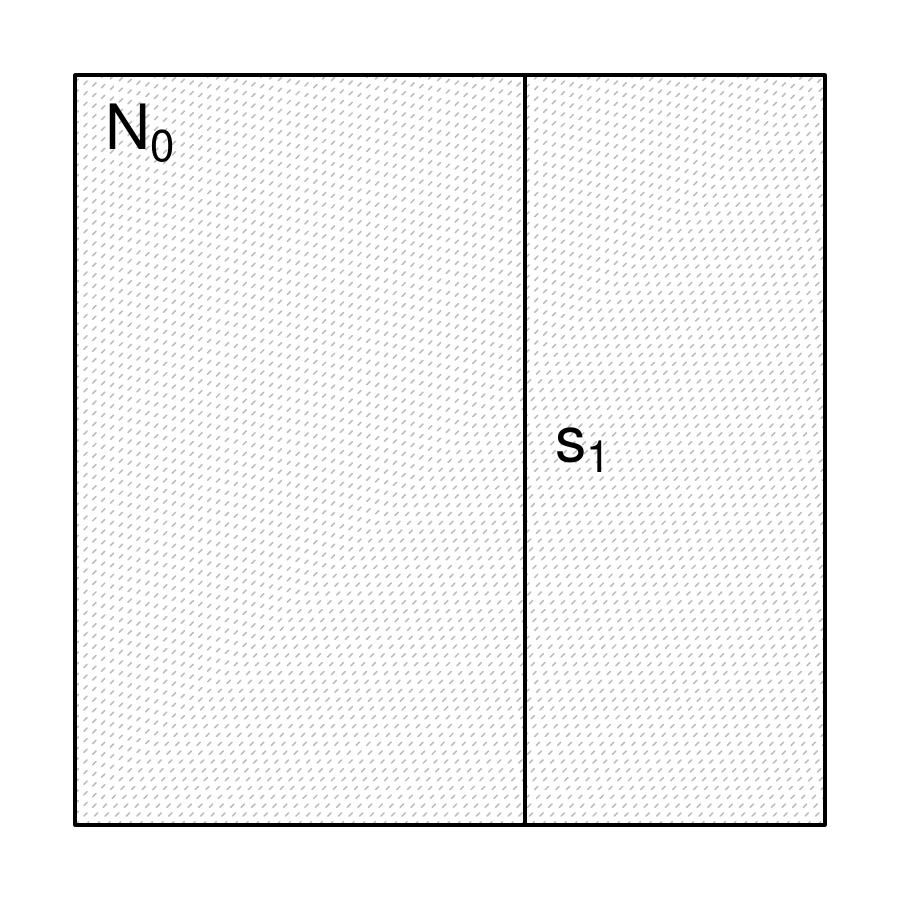}
  \end{subfigure}
  \begin{subfigure}{.32\textwidth}
    \centering
    \includegraphics[width=\linewidth]{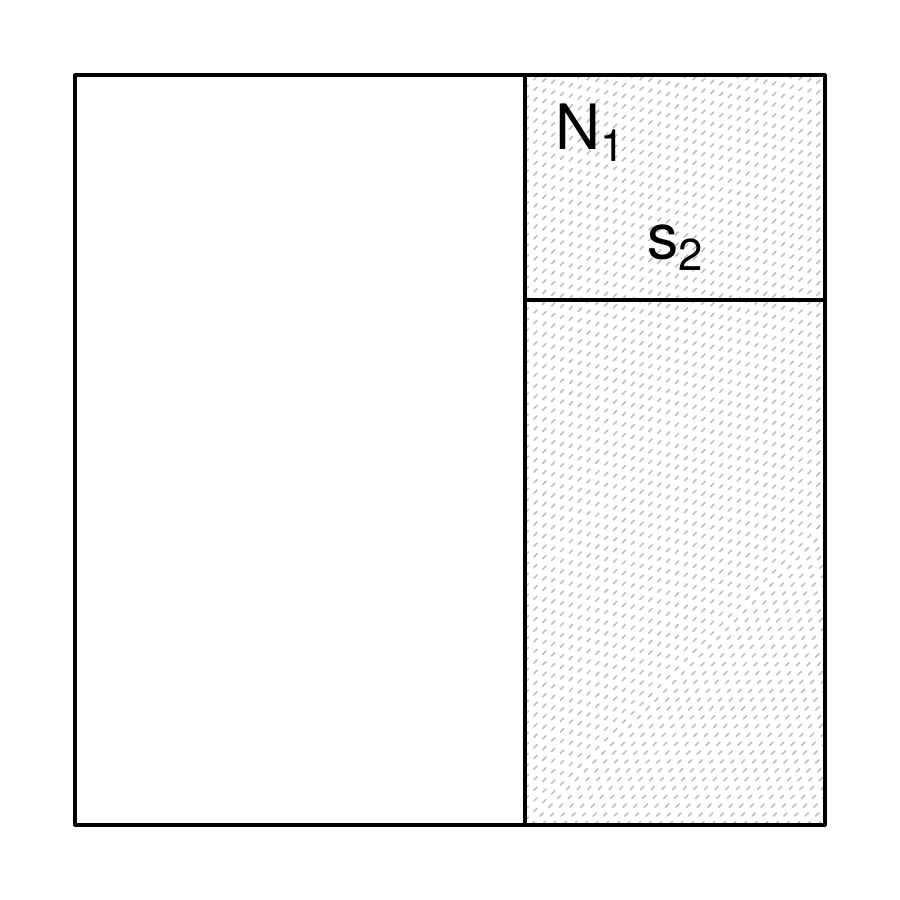}
  \end{subfigure}
  \begin{subfigure}{.32\textwidth}
    \centering
    \includegraphics[width=\linewidth]{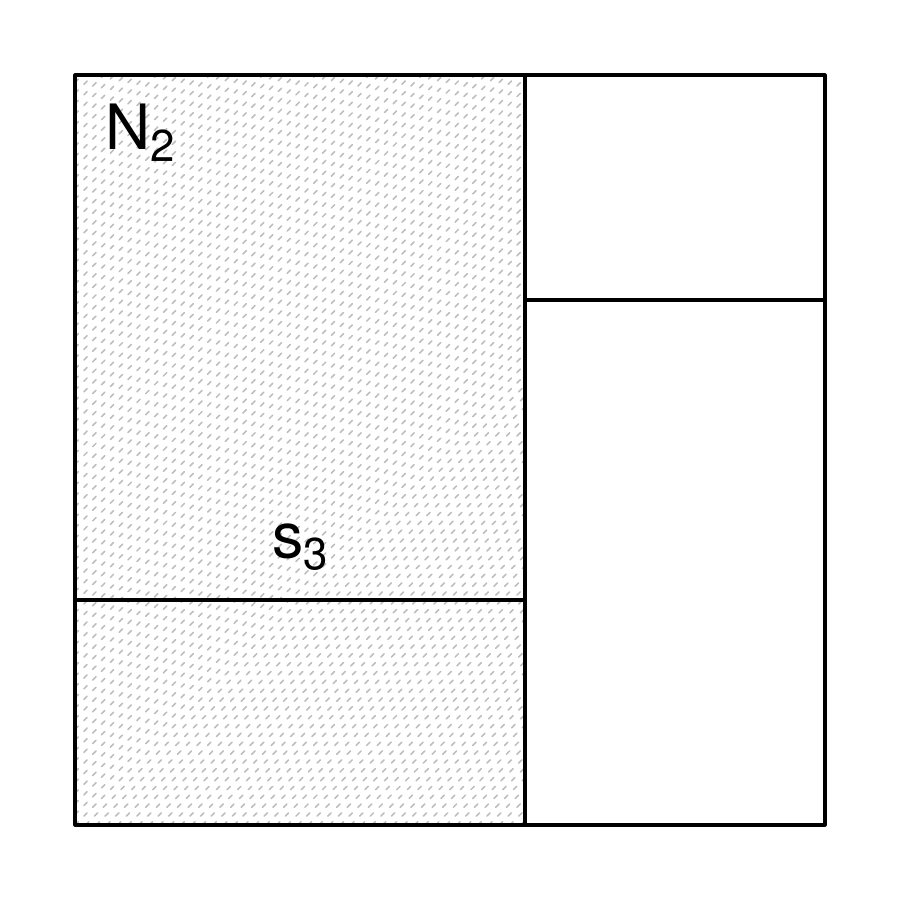}
  \end{subfigure}
  \caption{Regression tree notation. The shaded region in each panel is
  $N_0$, $N_1$, and $N_2$ from left to right. 
  The newly added line segments are $s_1$, $s_2$, and $s_3$.}
  \label{fig:tree_notation}
\end{figure}

To describe our algorithms, we introduce a set of
definitions for regression trees. These definitions are also 
conveyed in \cref{fig:tree_notation}. A \textit{node} $N$ is any set $N \subset \reals^d$. 
A \textit{split-point} $s = (j, l) \in \{1, 2, \ldots, d\} \times \reals$ is a tuple that 
indicates on which variable $j$ a split is to be made and at which location $l$.
For a fixed data set $\{(x_i, y_i)\}_{i=1}^n$, a \textit{regression tree} $T_K$ with
$K+1$ terminal nodes is an ordered sequence alternating between nodes and 
split-points $(N_0, s_1, N_1, s_2, \ldots, N_{K-1}, s_K)$. 
This notation indicates that the construction of the tree is assumed to have a well-defined history:
one starts with the tree $N_0 = \reals^d$ and then chooses a split-point $s_1 = (j_1, l_1)$ within $N_0$
that minimizes the squared-error loss. This split partitions $N_0$ into two regions, one of which
is chosen as the candidate for the next split. The next node to be split is $N_1$ and 
the split-point within $N_1$ is denoted $s_2$. After this second split, there are three
terminal nodes in the tree, one of which is selected as the candidate $N_2$ for the third split. 
The process repeats until a termination rule is reached. At the end, a $(K+1)$-element partition
of $\reals^d$ is obtained. 
In standard regression tree construction algorithms, the nodes are rectangular 
regions with faces aligned according to the basis vectors in $\reals^d$.

\section{Alpha-trimmed random forests}
\label{sec:alpha-trimming}
In this section we introduce a way to evaluate the utility of regression tree splits
and show how this evaluation can be used to prune trees adaptively and locally. 
Our methods are based on a modified Bayesian information criterion (BIC).
We start by showing how to prune an individual regression tree in \cref{sec:AI_pruning} 
using a generic information-based approach and then extend this method to random forests 
in \cref{sec:alpha-trimming_sub}.
In \cref{sec:possible_ICs} we discuss our modified Bayesian information 
criterion that can be used for regression tree pruning, and in 
\cref{sec:theory} we present some theoretical results.

\subsection{Accumulated information pruning algorithm}
\label{sec:AI_pruning}
Individual regression trees and ensembles of such trees can be pruned adaptively using 
our proposed \textit{accumulated information} pruning algorithm \citep{henrey2016statistical}. 
This approach \emph{does not}
rely on cross-validation, which is in contrast to the standard CART pruning algorithm of 
\cite{breiman1984classification}.
Our algorithm relies only on quantities that can be easily stored during tree 
construction. As a consequence, only one backwards pass is needed through the 
nodes of a tree in order to determine the structure of the pruned tree.

\subsubsection{Tree roots and stumps}
Starting from a fully-grown regression tree, the building blocks of the 
accumulated information pruning algorithm are \textit{tree roots} (one 
terminal node) and \textit{tree stumps} (two terminal nodes). 
We refer to these two types of simple trees (roots and stumps) as \textit{tree models}, 
because they offer a description for a possible data-generating mechanism for  
$(X_1, Y_1), \ldots, (X_n, Y_n)$.
The tree-building algorithm consists of a recursive application of a procedure 
that splits a tree root into two tree stumps. Evaluating the utility of a 
split can be reduced to comparing the information contained in 
statistical models representing the root and the stumps. 
Once an information quantity is 
defined for these two models, an iterative procedure that we refer to as 
accumulated information is used to assign information values to more complex tree structures. 

The tree root model is 
\[
  \label{eq:tree_root}  
  Y_i \mid X_i = x_i \sim N(\mu, \sigma_0^2),
\]
for some values of $\mu \in \reals$ and $\sigma_0^2 \geq 0$. 
Let $\mathbbm{1}(A)$ denote the indicator function on a set $A$ and $x^{(j)}$ 
the $j$-th coordinate of $x$. 
Suppose that this root is split into stumps according to whether $x^{(j)}$ is 
less than some value $l$. Then, the tree stump model is
\[
  \label{eq:tree_stump}
    Y_i \mid X_i = x_i &\sim N(m(x_i; j, l), \sigma_1^2), \\
    m(x_i; j, l)       &= \mu_1 \mathbbm{1}(x_i^{(j)} < l) + \mu_2 \mathbbm{1}(x_i^{(j)} 
    \geq l),
\]
for some $j \in \{1, 2, \ldots, d\}$, $l, \mu_1, \mu_2 \in \reals$, and $\sigma_1^2 \geq 0$.
Given a set of observed data $(x_i, y_i)$, $i=1,\ldots,n$, the log-likelihood for the 
tree root model is
\[
  \label{eq:tree_root_loglik}
    \ell_{0, n}(\mu, \sigma_0^2) 
    &\propto -\frac{n}{2} \log(2\pi\sigma_0^2) - \frac{1}{2\sigma_0^2} \sum_{i=1}^n (y_i - 
    \mu)^2,
\]
where we write $\propto$ to denote equivalence up to additive terms that are not 
functions of the parameters when working with the log-likelihood. 
The tree stump log-likelihood is 
\[
    \label{eq:tree_stump_loglik} 
    \ell_{1, n}(\mu_1, \mu_2, \sigma_1^2, j, l)
    \propto -\frac{n}{2} \log(2\pi\sigma_1^2) - \frac{1}{2\sigma_1^2} \sum_{i=1}^n (y_i - 
      m(x_i; j, l))^2.
\]

A common approach to comparing two models like these is to compute an information 
criterion on each model and determine which value is smaller, and hence the 
better-fitting model \citep{schwarz1978estimating}.
Common information criteria are of the form: $-2 \times (\text{maximized log-likelihood}) + (\text{penalty})$, 
where the penalty is related to the number of estimated model parameters. 
In \cref{sec:possible_ICs} we provide a discussion on 
appropriate penalties for the tree model. For now, we denote 
generic penalties for the tree root and tree stump models with $n$ data points 
as $P_{0, n}$ and $P_{1, n}$, respectively.

\subsubsection{Climbing up the tree}

\begin{algorithm}[t!]
\footnotesize
\caption{Accumulated information pruning}
\label{alg:accumulated_information}
\begin{algorithmic}
\Require Regression tree $T_K = (N_0, s_1, \ldots, N_{K-1}, s_K)$, 
  penalty functions $P_{0,n}$, $P_{1,n}$.
\State $T' \gets T_K$
\For{$k = 1, 2, \ldots, K$}
  \State $N, \, s \gets N_{K-k}, \, s_{K-k+1}$
    \Comment{obtain node and split point (traversing backwards)}
  \State $N_L, \, N_R \gets \texttt{children}(T', N, s)$
  \State $I_N \gets \texttt{parent\_information\_value}(T', N)$ 
    \Comment{information value of parent from \eqref{eq:information_parent}}
  \State $I_{N_L}, \, I_{N_R} \gets \texttt{child\_information\_value}(T', N_L, N_R)$
    \Comment{information value of children}
  \If{$I_N + P_{0, \abs{N}} \leq I_{N_L} + I_{N_R} + P_{1, \abs{N}}$}
    \State $T' \gets \texttt{merge\_nodes}(T', N, s)$ 
      \Comment{merge all descendents of node $N$}
  \EndIf 
\EndFor
\State \textbf{Return:} pruned regression tree $T'$
\end{algorithmic}
\end{algorithm}

Accumulated information pruning (\cref{alg:accumulated_information})
assumes that a regression tree has been fully grown up to some stopping criterion. 
It then starts from the bottom of the tree, at the leaf nodes, and attaches an information value 
to each node $N$ in the tree.
At the beginning of the pruning algorithm,
the information value of each node in the tree is undefined. 
Suppose that the tree contains $K+1$ terminal nodes and was 
constructed by making $K$ splits on the nodes $N_0, N_1, \ldots, N_{K-1}$ in this exact order. 
We prune the tree by traversing the nodes in reverse order, starting with node $N_{K-1}$.

\bigskip \noindent \textbf{Parent nodes} \\
Consider a parent node $N$ and its two children $N_L$ and $N_R$. 
At the beginning of the pruning 
algorithm we have $N = N_{K-1}$ and and $s=s_K=(j_K ,l_K)$, using the notation of \cref{sec:preliminaries}.
Then, $N_L = \cbra{x \in N : x^{(j_K)} < l_K}$ and $N_R = \cbra{x \in N : x^{(j_K)} \geq l_K}$.

A decision is to be made whether to merge the two nodes into a single new 
terminal node $N$ or to keep the split between $N_L$ and $N_R$. Using maximum likelihood estimation for parameters, let $\abs{N}$ 
be the number of observations in node $N$ and define the estimated mean and 
variance under the tree root model as
\[
  \label{eq:mean_var_tree_root}
  \hat{\mu} = \bar{y}_N = \frac{1}{\abs{N}} \sum_{i=1}^n y_i \mathbbm{1}(x_i \in N), \qquad 
  \hat{\sigma}_0^2 = \frac{1}{\abs{N}} \sum_{i=1}^n (y_i - \bar{y}_N)^2 \mathbbm{1}(x_i \in N).
\]
We assign the information value \emph{before adding an appropriate penalty} 
in the parent node $N$ as
\[
  \label{eq:information_parent}
  I_N 
  &\leftarrow \abs{N} \log(2\pi\hat{\sigma}_0^2) + \abs{N}.
\]
From here we consider two cases for the child nodes before we consider merging them:
both $I_{N_L}$ and $I_{N_R}$ are undefined or only one of them is undefined.

\bigskip \noindent \textbf{Child nodes: case 1} \\
If the information value in \textit{both nodes} $N_L$ and $N_R$ is undefined
(i.e., they are both leaf nodes and have no children), we define the 
quantities $I_{N_L}$ and $I_{N_R}$ as follows. 
Define the respective estimated means under the tree stump model as 
$\hat{\mu}_1 = \bar{y}_{N_L}$, $\hat{\mu}_2 = \bar{y}_{N_R}$, and the estimated variance 
under the tree stump model as
\[
  \hat{\sigma}_1^2 
  = \frac{1}{\abs{N}} \sbra{\sum_{i=1}^n (y_i - \hat{\mu}_1)^2 \mathbbm{1}(x_i \in N_L) 
    + \sum_{i=1}^n (y_i - \hat{\mu}_2)^2 \mathbbm{1}(x_i \in N_R)}.
\]
The information values in nodes $N_L$ and $N_R$ \emph{before adding a penalty} are then
\[
  \label{eq:information_children_undef}
  I_{N_L} &\leftarrow \abs{N_L} \log(2\pi\hat{\sigma}_1^2) + 
    \frac{1}{\hat{\sigma}_1^2} \sum_{i=1}^n (y_i - \hat{\mu}_1)^2 \mathbbm{1}(x_i \in N_L) \\
  I_{N_R} &\leftarrow \abs{N_R} \log(2\pi\hat{\sigma}_1^2) + 
    \frac{1}{\hat{\sigma}_1^2} \sum_{i=1}^n (y_i - \hat{\mu}_2)^2 \mathbbm{1}(x_i \in N_R).
\]

\bigskip \noindent \textbf{Child nodes: case 2} \\
In the case that only one of $I_{N_L}$ or $I_{N_R}$ is undefined, 
the same general formula is used to compute the information value in the 
undefined node, but with a different estimate of $\sigma^2$. Suppose 
that $I_{N_L}$ is undefined and $I_{N_R}$ is already defined. 
The estimate of $\sigma^2$ used to 
define $I_{N_L}$ is the sum of squared errors in node $N_L$ plus the sum of squared errors in all 
remaining terminal nodes of the subtree rooted at node $N_R$, all divided by 
$\abs{N} = \abs{N_L} + \abs{N_R}$.

\bigskip \noindent \textbf{Merging the child nodes and accumulating information} \\
Two nodes are merged if the information criterion computed in the root is smaller 
than in the stumps, indicating that the model with a single mean is better. 
Formally, the nodes $N_L$ and $N_R$ are merged if
\[
  \label{eq:merge_criterion}
  I_N + P_{0, \abs{N}} \leq I_{N_L} + I_{N_R} + P_{1, \abs{N}},
\]
where $P_{0, \abs{N}}, P_{1, \abs{N}} \geq 0$ are penalties for the tree root and stump models 
with $\abs{N}$ observations, which we specify in \cref{sec:possible_ICs}.
In the case of merging the two children of node $N$, 
it becomes a new terminal node, and $I_N$ is set to be undefined again.
If the condition given by \eqref{eq:merge_criterion} is not met, 
the regression tree split is kept. In this case, the information value of node $N$ is updated to
\[ 
  I_N \leftarrow I_{N_L} + I_{N_R} + P_{1, \abs{N}} - P_{0, \abs{N}},
\]
so that the root reflects the information contained in the nodes beneath it. 

We note that it is still possible that the nodes $N_L$ and $N_R$ will be merged at a later point along with 
other nodes in an entire branch as the pruning algorithm climbs up the tree. The accumulated 
information algorithm continues to prune by considering the next node in the reversed tree 
construction sequence. For example, if we had just considered node $N = N_{K-1}$, 
then the next node under consideration would be $N_{K-2}$.
The accumulated information process is repeated until every node in the tree has been assigned 
an information value. The last node to be assigned an information value is always the root node of the tree.
Once the root node of the tree is reached, the procedure terminates.
The full procedure is summarized in \cref{alg:accumulated_information}.

\subsection{Alpha-trimming}
\label{sec:alpha-trimming_sub}
The penalties used in the accumulated information pruning algorithm for a single 
tree are designed to produce a regression machine that balances the effects of 
bias and variance. However, in a random forest ensemble, we prefer trees that 
have little bias, even at the cost of extra variance. Thus, we need to adjust 
the penalties to make them more suitable for ensemble learning.
To achieve this, we develop the alpha-trimming procedure 
(\cref{alg:alpha_trimming}) for random forests.
To control the amount of pruning, we introduce an additional tuning 
parameter, $\alpha \geq 0$,
and we amend the node-merging step from \cref{sec:AI_pruning} as follows. 
Two nodes $N_L$ and $N_R$ are merged into node $N$ if 
\[
  \label{eq:merge_criterion_alpha}
  I_N + \alpha P_{0, \abs{N}} \leq I_{N_L} + I_{N_R} + \alpha P_{1, \abs{N}}.
\]
If the split between $N_L$ and $N_R$ is kept, $I_N$ is updated to
\[
  I_N \leftarrow I_{N_L} + I_{N_R} + \alpha (P_{1, \abs{N}} - P_{0, \abs{N}}). 
\] 
This procedure is repeated for a given value of $\alpha$ for each tree in the random forest 
independently, which can be easily parallelized across trees. 

The parameter $\alpha$ serves 
to balance the bias-variance trade-off in the individual trees with 
consideration for their use in the ensemble.  
Note that if $\alpha=1$, each tree is pruned to be a good standalone predictor.
If $\alpha=0$, then no pruning is performed and the standard random forest is obtained. 
Therefore, alpha-trimming contains the standard random forest as a special case when $\alpha=0$. 
We note that in some cases with a considerably low SNR it 
is favourable to try values of $\alpha \geq 1$. 
In our experience, an optimal value of $\alpha$ is often between $0$ and $3$.

\begin{algorithm}[t!]
    \footnotesize
    \caption{Alpha-trimming}
    \label{alg:alpha_trimming}
    \begin{algorithmic}[1]
    \Require Random forest with $B$ trees $\cbra{T_{K_b, b}}_{b=1}^B$, 
      grid of $\alpha$ values $\mcA$, penalty functions $P_{0,n}$, $P_{1,n}$
    \For{$\alpha \in \mcA$}
      \For{$b = 1,2,\ldots,B$} \label{line:loop_AI_pruning}
        \State $T_{K_b, b}^\alpha \gets 
          \texttt{accumulated\_information}(T_{K_b, b}, \, \alpha P_{0,n}, \, \alpha P_{1,n})$ 
          \Comment{prune each tree with \cref{alg:accumulated_information}}
      \EndFor
    \EndFor 
    \State $\alpha^\star \gets \argmin_{\alpha \in \mcA} \texttt{OOB}(T_{K_b, b}^\alpha)$ 
      \label{line:OOB}
      \Comment{select $\alpha$ based on out-of-bag error estimates}
    \State \textbf{Return:} optimal alpha-trimmed random forest $\cbra{T_{K_b, b}^{\alpha^\star}}$
    \end{algorithmic}
\end{algorithm}

One major advantage of our method is that cross-validation is not required.
We can choose a grid of values for $\alpha$ and use the out-of-bag (OOB) 
observations from each of the trees---those left out of the bootstrap resample 
for the tree---to obtain an estimate of the mean squared 
prediction error for a given value of $\alpha$. 

For a fixed tree size with $K+1$ terminal nodes, 
accumulated information pruning for a fixed value of $\alpha$ 
has a theoretical computational complexity of $O(K)$ (see \cref{prop:comp_complexity}). 
Note that this is independent of the data set size $n$ as the data set size grows,
because the sums of residuals required for the information values can be stored 
during tree construction.
In contrast, tree construction time in its most basic form 
depends on the data set size $n$, the dimension $d$, and 
the number of tree splits $K$. The advantage of alpha-trimming 
(with accumulated information pruning) is that it \textit{does not} require repeated 
tree construction nor cross-validation while still being able to prune adaptively.

\subsection{Proposed information criterion}
\label{sec:possible_ICs}
A convenient property of the proposed pruning approach is that it only requires an information 
quantity to be defined for the tree root and tree stump models. Information 
quantities are then recursively extended to larger tree structures. 
In this section we introduce a modified Bayesian information criterion (BIC) that 
can be used for pruning regression trees and that possesses a desirable consistency 
property for choosing between tree root and tree stump models \citep{surjanovic2021pruning}. 
Our consistency result is related to the modified BIC change-point detection 
results presented by \cite{shen2011developing}.

For the tree root model given by \eqref{eq:tree_root}, which does not contain any
split-point parameters, we use the standard BIC. That is, 
we add the penalty $P_{0,n} = 2 \log(n)$ 
to $(-2) \times (\text{tree root log-likelihood})$ in \eqref{eq:tree_root_loglik} 
in order to account for the two estimated parameters: $\mu$ and $\sigma^2$.
For the tree stump model given by \eqref{eq:tree_stump} we use a modified BIC 
similar to the information criteria for change-point detection models 
\citep{ninomiya2015change,shen2011developing}. 
The penalty in this case is $P_{1, n} = 5 \log(n)$. Note that in this case we 
estimate the parameters $\mu_1$, $\mu_2$, $\sigma^2$, and $s$.

\subsection{Theoretical results}
\label{sec:theory}
We prove that our modified BIC is consistent for choosing between the tree root and stump
models (\cref{prop:BIC_consistency_1,prop:BIC_consistency_2}).
We also show that for a model of the regression tree construction algorithm,
the amount of pruning that should be performed in order to minimize the mean squared 
prediction error is a function of the signal-to-noise ratio (\cref{prop:SNR_MSPE}), 
justifying our study of the SNR as a criterion for the extent to which 
regression trees are pruned.
Finally, we establish the computational complexity of fitting alpha-trimmed RFs 
(\cref{prop:comp_complexity}).
The proofs of theoretical results can be found in the appendices.

To state our results, we first provide a more formal definition of 
our proposed BIC for the tree root and stump models, following the presentation in 
\cite{surjanovic2021pruning}.
Consider $Z_i = (X_i, Y_i)$ i.i.d.~for $i = 1, 2,\ldots, n$. 
For simplicity we suppose that the distribution 
of each $X_i$ is supported on $[0,1]^d$, although the definitions can be extended 
to other domains.

For the tree root model, if $\sigma^2$ is \textit{known}, we define
\[
  \text{BIC}_0(Z_1, \ldots, Z_n; \sigma^2) 
  = n \log(2 \pi \sigma^2) + 
    \frac{1}{\sigma^2} \sum_{i=1}^n (Y_i - \hat{\mu})^2 + \log(n).
\]
If $\sigma^2$ is \textit{unknown} and we use the maximum-likelihood estimate $\hat{\sigma}^2_0$, set
\[
  \text{BIC}_0(Z_1, \ldots, Z_n; \hat{\sigma}^2_0) 
  = n \log(2 \pi \hat{\sigma}^2_0) + \frac{1}{\hat{\sigma}^2_0} 
    \sum_{i=1}^n (Y_i - \hat{\mu})^2 + 2 \log(n).
\]

For the tree stump model we assume that the split-point location is estimated by the CART 
tree construction algorithm with a guarantee that each node contains at least 
one element after performing the split. 
Let $Z_1(j), \ldots, Z_n(j)$ for $j \in \cbra{1,2,\ldots, d}$ 
be the permutation of $Z_1, \ldots, Z_n$ ordered according to the 
$j$-th coordinate of the set of random variables $\cbra{X_i}_{i=1}^n$ 
so that they are sorted in increasing value.
Note that this permutation is a random permutation (i.e., it depends on the realized 
values of the $\cbra{X_i}_{i=1}^n$).\footnote{More formally, 
if $(\Omega, \mathcal{B}, \mathbb{P})$ is the underlying 
probability space, then for any $\omega \in \Omega$ we have 
$X_1(j)(\omega) \leq X_2(j)(\omega) \leq \ldots \leq X_n(j)(\omega)$.}
For $n_L \in \cbra{1, \ldots, n-1}$ define $n_R = n - n_L$ and 
\[ 
  \bar{Y}_{L, n_L}(j) = n_L^{-1} \rbra{Y_1(j) + \ldots + Y_{n_L}(j)}, \qquad
  \bar{Y}_{R, n_L}(j) = n_R^{-1} \rbra{Y_{n_L+1}(j) + \ldots + Y_n(j)}.
\]
Finally, set
\[
  &\text{BIC}_1(Z_1, \ldots, Z_n; \hat{\sigma}^2_1) \\
  &{\quad} = n \log(2\pi \hat{\sigma}^2_1) + 
    \min_{\substack{n_L = 1, \ldots, n-1 \\ j=1,\ldots, d}} \frac{1}{\hat{\sigma}^2_1} 
    \sbra{ \sum_{i=1}^{n_L} (Y_i(j) - \bar{Y}_{L,n_L}(j))^2 + 
      \sum_{i=n_L + 1}^n (Y_i(j) - \bar{Y}_{R,n_L}(j))^2 } \\ 
    &{\qquad} + 5 \log(n).
\]
If $\sigma^2$ is known we use a penalty of $4 \log(n)$.

\cref{prop:BIC_consistency_1,prop:BIC_consistency_2} show that our 
modified BIC correctly chooses between the tree root or tree stump models with 
probability approaching one as the sample size tends to infinity, provided that one of 
the two models is correct and a consistent estimator of the conditional variance 
is used. 

\bprop
\label{prop:BIC_consistency_1}
Suppose that $Z_i = (X_i, Y_i)$ are i.i.d.~with the distribution of each $X_i$
supported on $[0,1]^d$ with a strictly positive density with 
respect to Lebesgue measure on $[0,1]^d$. 
Further, suppose that $Y_i \mid X_i = x_i \sim \distNorm(\mu, \sigma^2)$ for some 
$\mu \in \reals$ and $\sigma^2_0 > 0$; i.e., the true model is the tree root model.
If $\hat{\sigma}^2_n \xrightarrow{p} \sigma^2_0$, then 
\[
  \Pr(\text{BIC}_0(Z_1, \ldots, Z_n; \hat{\sigma}^2_n) < 
    \text{BIC}_1(Z_1, \ldots, Z_n); \hat{\sigma}^2_n) 
  \to 1, \quad n \to \infty.
\]
\eprop 

\bprop
\label{prop:BIC_consistency_2}
Consider $Z_i = (X_i, Y_i)$ i.i.d.~with the same distributional assumptions on each $X_i$ 
as in \cref{prop:BIC_consistency_1}. 
Suppose that $Y_i \mid X_i = x_i \sim \distNorm(m(x_i), \sigma^2_1)$ where 
$\sigma^2_1 > 0$ and 
$m(x) = \mu_1 \mathbbm{1}(x^{(j)} < l) + \mu_2 \mathbbm{1}(x^{(j)} \geq l)$
for some $\mu_1, \mu_2 \in \reals$ with $\mu_1 \neq \mu_2$, $j \in \cbra{1,2,\ldots,d}$, and $l \in (0,1)$; i.e., the true model is the tree stump model. 
If $\hat{\sigma}^2_n \xrightarrow{p} \sigma^2_1$, then  
\[
  \Pr(\text{BIC}_1(Z_1, \ldots, Z_n; \hat{\sigma}^2_n) < 
    \text{BIC}_0(Z_1, \ldots, Z_n; \hat{\sigma}^2_n)) 
  \to 1, \quad n \to \infty.
\]
\eprop

We also study the relationship between the number of regression tree split-points, the SNR, 
and the mean squared prediction error. To obtain closed-form results, 
we consider a simplified model of the regression 
tree algorithm. 
We assume that the same number of data points are included within 
each region of the partitioned covariate space and that our estimated regression 
curve is the average of the responses within each region.

\bprop 
\label{prop:SNR_MSPE}
Suppose for $i=1,2,\ldots, n$ that 
$Y_i \mid X_i = x_i \sim \distNorm(\beta x_i, \sigma^2)$ for some $\beta \in \reals$ 
and $\sigma^2 > 0$ so that the true regression curve is $\mu(x) = \beta x$. 
Assume that $n=mk$ for some $m, k \in \nats$ and that 
the distributions of the random variables $\cbra{X_i}_{i=1}^{km}$ are such that 
\[
  X_1, \ldots, X_m &\sim U([0, 1/k)) \\
  X_{m+1}, \ldots, X_{2m} &\sim U([1/k, 2/k)) \\
  &\vdots \\
  X_{(k-1)m+1}, \ldots, X_{km} &\sim U([(k-1)/k, 1]).
\]
Define the regression curve estimate $\hat{\mu}(x)$ for $x \in [0,1]$ to be 
a piecewise-constant function such that if $x \in [j/k, (j+1)/k)$ for some 
$j \in \cbra{0, 1, \ldots, k-1}$ then 
\[
  \hat{\mu}(x) 
  = m^{-1} \rbra{Y_{jm + 1} + \cdots Y_{(j+1)m}}.
\]
Then, for any fixed $j \in \cbra{0, 1, \ldots, k-1}$,
\[
  \label{eq:MSPE_model}
  \EE[(\hat{\mu}(X) - \mu(X))^2] 
  = \frac{\sigma^2}{m} + \rbra{1 + \frac{1}{m}} \cdot \frac{\beta^2}{12 k^2}, 
  \qquad X \sim U([j/k, (j+1)/k]).
\]
\eprop 

We now show how this result can be used to suggest that when the SNR is high we should grow deeper trees.
Informally, if we suppose that the result above holds even if $m = n/k$ 
is not a positive integer, then \eqref{eq:MSPE_model} for a given $n$ is minimized
as a function of $k$ by solving the equation 
\[
  \label{eq:MSPE_minimizer}
  \rbra{\frac{\sigma^2}{\beta^2 n}} k^3 - \rbra{\frac{1}{12n}} k - \frac{1}{6} = 0.
\]
Note that the solution to \eqref{eq:MSPE_minimizer}
is necessarily only a function of the quantity $\gamma := \abs{\beta/\sigma}$, 
which is the SNR. For fixed $\gamma \neq 0$, we have 
$k \sim (\gamma^2 n/6)^{1/3}$ as $n \to \infty$. 

Finally, we assess the computational complexity of constructing an alpha-trimmed RF, 
which \emph{does not} require cross-validation to tune the trimming parameter $\alpha$.
Additionally, the computational complexity of the alpha-trimming procedure does not depend on the 
dimension $d$, and alpha-trimming can be easily performed in parallel over the number of trees 
$B$ and the number of tuning parameters $\abs{\mcA}$.

\bprop 
\label{prop:comp_complexity} 
Consider a forest of $B$ trees $\cbra{T_{K_b,b}}_{b=1}^B$, where the $b^\text{th}$ 
tree has $K_b+1$ terminal nodes. Provided that each $K_b = O(n)$, where $n$ is the number 
of data points, and that the trees are balanced, we have the following:
\begin{enumerate}
  \item Building a single fully-grown regression tree has a complexity of 
  $O(d n \log_2 n)$ and the cost of building a forest is $O(B d n \log_2 n)$.
  \item Pruning a single regression tree with accumulated information (\cref{alg:accumulated_information})
  can be done in $O(n)$ time, and therefore the cost of alpha-trimming a RF 
  for a fixed value of $\alpha$ is $O(B n)$. 
  \item Evaluating OOB error for a forest is $O(B n \log_2 n)$ and therefore finding 
  an optimal value of $\alpha$ over a grid of size $\abs{\mcA}$ can be done in 
  $O(\abs{\mcA} B n \log_2 n)$ time.
\end{enumerate} 
\eprop

\subsection{Previous work}
\label{sec:previous_work}

Previous authors have identified that controlling the tree depth in a random forest 
can be beneficial in terms of predictive accuracy 
\citep{segal2004machine,zhou2019pruning,duroux2018rf,lin2006random}.
The general consensus among these authors is that the tree depth is a potentially 
valuable parameter that should be tuned. \cite{zhou2019pruning} emphasize that 
``shallow trees are advantageous when the signal-to-noise ratio in the data is low,'' 
which is similar to the conclusion drawn in our work. However, our approach differs 
from that of \cite{zhou2019pruning} in that we consider approaches to pruning 
regression trees in settings where the signal-to-noise ratio may vary across the domain.
Other work on controlling the depth of trees in a random forest, such as by 
\cite{liu2023forestprune}, has focused more on the computational and storage gains 
that can be obtained by reducing the overall size of the ensemble. 

In this work we consider pruning each tree within a RF ensemble by removing non-informative branches 
with the goal of offering better predictive accuracy.
The closest work to ours in this regard is by \cite{jiang2017forest}, who focus on the 
problem of pruning classification trees within a RF ensemble using an information-based approach. 
Their Algorithm 1 is similar to our proposed pruning approach, except that they do not 
consider tuning the information penalty to find one that is optimal for predictive performance. 
\cite{jiang2017forest} focus on the classification setting and do not comment on 
when pruning would be beneficial (e.g., in low SNR settings).

Other work on the pruning of random forests has focused on the removal of entire 
trees from the ensemble. For instance, \cite{manzali2023rfpruning} offer a recent review 
of this literature. 
While this line of work can again be useful for reducing the memory requirements of 
ensembles such as RFs, the removal of entire trees does not directly address 
the need to decrease the depth of trees when the SNR is low. 
Furthermore, such an approach to pruning of ensembles does not consider the setting 
where the SNR varies across the domain.

With respect to individual trees, choosing splits and pruning using a 
likelihood-based approach has been previously explored. 
\cite{ciampi1991generalized} propose pruning trees by considering the 
set of all possible subtrees starting from the root (top) node. They suggest to compute an AIC 
on each of these subtrees and select the optimally pruned tree with the smallest 
Akaike information criterion (AIC). 
\cite{su2004maximum} similarly frame regression trees in a maximum likelihood context and 
develop a different pruning algorithm. 
However, both of these approaches suffer from a similar flaw. They ignore the fact that the split-point 
location is an unknown parameter at each step that must be estimated from the data. The 
pruning algorithms of \cite{ciampi1991generalized} and \cite{su2004maximum} treat these 
parameter values as if they are known in advance. \cite{su2004maximum} acknowledge that 
the split-point location is an estimated parameter, but they do not consider it in their penalty calculations. 
In contrast, in our work we directly address the estimation of the split-point parameter 
in our information criterion.

\section{Simulation study design}
\label{sec:sim_design}
We conduct a simulation study to examine the performance of: 
(1) our alpha-trimmed RFs (\texttt{AlphaTrim}); 
(2) default RFs (\texttt{RF}) with a fixed minimum node size of 5 for further splitting; and 
(3) standard RFs tuned by trying various node sizes (\texttt{RF\_tuned}). 
The \texttt{RF\_tuned} procedure differs from \texttt{AlphaTrim} because adjusting
the minimum node size globally shrinks or expands the tree rather than adaptively selecting
regions to prune based on the local SNR.
As noted previously, tuning the minimum node size is seldom recommended in practice
and so it is of general interest whether any method of controlling tree size can 
improve the predictive performance in a random forest.

The first part of the simulation study considers 
several basic examples that illustrate the behaviour of standard random forests as the 
SNR varies from low to high. In the second part of the simulation
study we examine the performance of the various RF methods on 42 data sets from 
the work of \cite{chipman2010bart}, along with four additional synthetic data sets. 
All simulations are performed in \textsf{R} on an Intel i9 CPU with 32 GB of RAM.

\subsection{SNR experiments}
\label{sec:SNR_experiments}
Standard \emph{fully-grown} RFs struggle to model surfaces that are flat (low SNR) 
or have a mixture of regions with low and high SNR. We demonstrate 
these issues with some simulated examples.
First, we fit a RF using the \texttt{ranger} package in \textsf{R}. In this package, the 
default is to control regression tree size by allowing splits on 
nodes with 5 or more observations. 
We denote this tuning parameter as $n_\text{min}$ so that with the 
\texttt{ranger} defaults we have $n_\text{min} = 5$, which is controlled by 
the argument \texttt{min.node.size}. 
Note that this does not preclude creating smaller nodes, including nodes of size 1. 
We then re-fit the RFs using values of 
$n_\text{min} \in \{10, 20, 50, 100, 200, 300, 400, 500\}$.
Using the training set OOB errors, we estimate which value of $n_\text{min}$ should be optimal 
and then make predictions on the test set, denoting the predictor $\texttt{RF-tuned}$.
In all cases we use $B = 750$ trees.
For each model and data set we compute the estimated root mean squared prediction error (RMSPE) using 
a test set with 1500 observations. 
We let $d'$ denote the size of the subset of variables selected at random for each tree 
split in the forest. This parameter is controlled by the argument \texttt{mtry} in 
and we use the default provided by that package for regression trees, which is equal to $d/3$. 
For the alpha-trimmed trees, we set $n_\text{min} = 3$ and then adjust 
$\alpha \in \cbra{0, 0.1, 0.2, \ldots, 3.0}$ to control the size of the trees.

\subsubsection{Constant SNR}
To demonstrate the difficulties that RFs have in low SNR problems, we generate 
three data sets. The data are drawn from the model
\[
    Y_i \mid X_i = x_i \sim \distNorm\rbra{ \beta \cdot \sum_{j=1}^5 x_{ij}, 1 }, \qquad
    X_i \stackrel{iid}{\sim} U([0, 1]^5),
\]
for $i=1,2,\ldots, n$, with $n=500$ in the training set and $n=1500$ in the test set.
We consider $\beta=0$ (no signal), $\beta=0.5$ (low SNR), and $\beta=3$ (high SNR).

\subsubsection{Mixed low and high SNR}
We also consider a situation that demonstrates why tuning the sizes of trees in a global
manner via parameters such as \texttt{min.node.size} is not always appropriate. 
All of the settings are the same as in the first example with the exception 
that the response is now
\[
  Y_i \mid X_i = x_i 
  \sim \distNorm\rbra{ 10 (x_i^{(1)} - 0.5) \cdot \mathbbm{1}(x_i^{(1)} \geq 0.5), 1 },
\]
for $i=1,2,\ldots,n$, with $n=500$ in the training set and $n=1500$ in the test set. 
This makes the true regression function take on an ``elbow'' shape with a joint at $x^{(1)}=0.5$. 
Although the model is simple, it results in mixed levels of SNR in different regions.

\subsection{Data sets and methods}
We examine the performance of alpha-trimmed RFs compared to standard and tuned RFs on 42 
data sets from \cite{chipman2010bart}, along with four additional data sets: 
\texttt{constant}, \texttt{elbow}, \texttt{logistic}, and \texttt{sine}. 
These additional data sets are artificially generated and are described in the appendices. 

The alpha-trimmed RF is tuned by considering values of $\alpha$ in 
$\{ 0, 0.1, 0.2, \ldots, 3.0\}$. All RFs are fit with $B=750$ trees. For the 
alpha-trimmed RF and the tuned RF we try values of 
$d' \in \{ 1, \floor{\sqrt{d}}, \floor{d/3}, \floor{2d/3}, d-1 \}$, excluding any 
values that are less than 1. For the tuned RF, instead of tuning $\alpha$ we 
tune $n_\text{min}$ by trying values in 
$\{\floor{2 (n/2)^0} = 2, \floor{2 \cdot (n/2)^{1/30}}, \floor{2 \cdot (n/2)^{2/30}}, 
\ldots, \floor{2 \cdot (n/2)^1} = n\}$.
The alpha-trimmed RF always uses a constant value of $n_\text{min} = 3$.
For each of the data sets we compute the estimated RMSPE using six-fold 
cross-validation. This process is repeated 10 times
using bootstrapped data sets  
so that an approximate 95\% $z$-based confidence interval for the (ratio of) RMSPEs is obtained. 

\section{Simulation results}
\label{sec:sim_results}

\subsection{SNR examples}

\begin{table}[t!]
\caption{Estimated RMSPEs from fitting an 
alpha-trimmed RF, RFs with varying restrictions on 
terminal node sizes, and a RF with tuned terminal node size.
The simulations are performed with the data sets described in \cref{sec:SNR_experiments}. 
Bolded entries in each column indicate methods with the lowest estimated RMSPE.}
\begin{center}
\label{tab:SNR_results}
\begin{tabular}{l|rrrr}
\hline
& \multicolumn{3}{c}{SNR}& \\ \cline {2-4}
Method              & 0              & Low            & High           & Elbow \\ \hline
\texttt{AlphaTrim}  & 0.975          & 1.030          & \textbf{1.227} & \textbf{1.069} \\ 
\texttt{RF-tuned}   & \textbf{0.974} & 1.025          & 1.228          & 1.079 \\ 
\texttt{RF-5}       & 1.016          & 1.053          & 1.228          & 1.077 \\ 
\texttt{RF-10}      & 1.013          & 1.046          & 1.235          & 1.079 \\ 
\texttt{RF-20}      & 1.006          & 1.039          & 1.265          & 1.078 \\ 
\texttt{RF-50}      & 0.998          & 1.030          & 1.377          & 1.081 \\ 
\texttt{RF-100}     & 0.992          & 1.025          & 1.530          & 1.104 \\ 
\texttt{RF-200}     & 0.986          & \textbf{1.023} & 1.735          & 1.170 \\ 
\texttt{RF-300}     & 0.983          & 1.028          & 1.870          & 1.225 \\ 
\texttt{RF-400}     & 0.979          & 1.034          & 1.962          & 1.279 \\ 
\texttt{RF-500}     & \textbf{0.974} & 1.058          & 2.218          & 1.876 \\ 

\hline
\end{tabular}
\end{center}
\end{table}

\cref{tab:SNR_results} shows the estimated RMSPEs of the various RFs. 
Here, \texttt{RF-x} is an untuned RF with $n_\text{min}$ equal to \texttt{x}. 
When the slope is zero, the best predictor for the constant regression function is 
the forest with the most shallow trees. 
The more splits the trees make, the more they overfit the training data and 
the worse the predictions are on average in terms of variance. 
When the SNR is low but non-zero, the RFs with the 
largest trees overfit, while those with the smallest trees seem to underfit and do 
not capture the signal adequately. 
Finally, in the high SNR case, the larger 
trees show their merit. The RFs with larger trees (smaller value of $n_\text{min}$) 
fit much better than RFs using smaller trees (larger value of $n_\text{min}$).
In all three cases, the alpha-trimmed RFs have an RMSPE comparable to the RF with tuned $n_\text{min}$. 
When the SNR has regions of both high and low SNR, the alpha-trimmed trees have the lowest 
estimated RMSPE.

\begin{figure}[t!]
    \centering
    \begin{subfigure}{.4\textwidth}
      \centering
      \includegraphics[width=\linewidth]{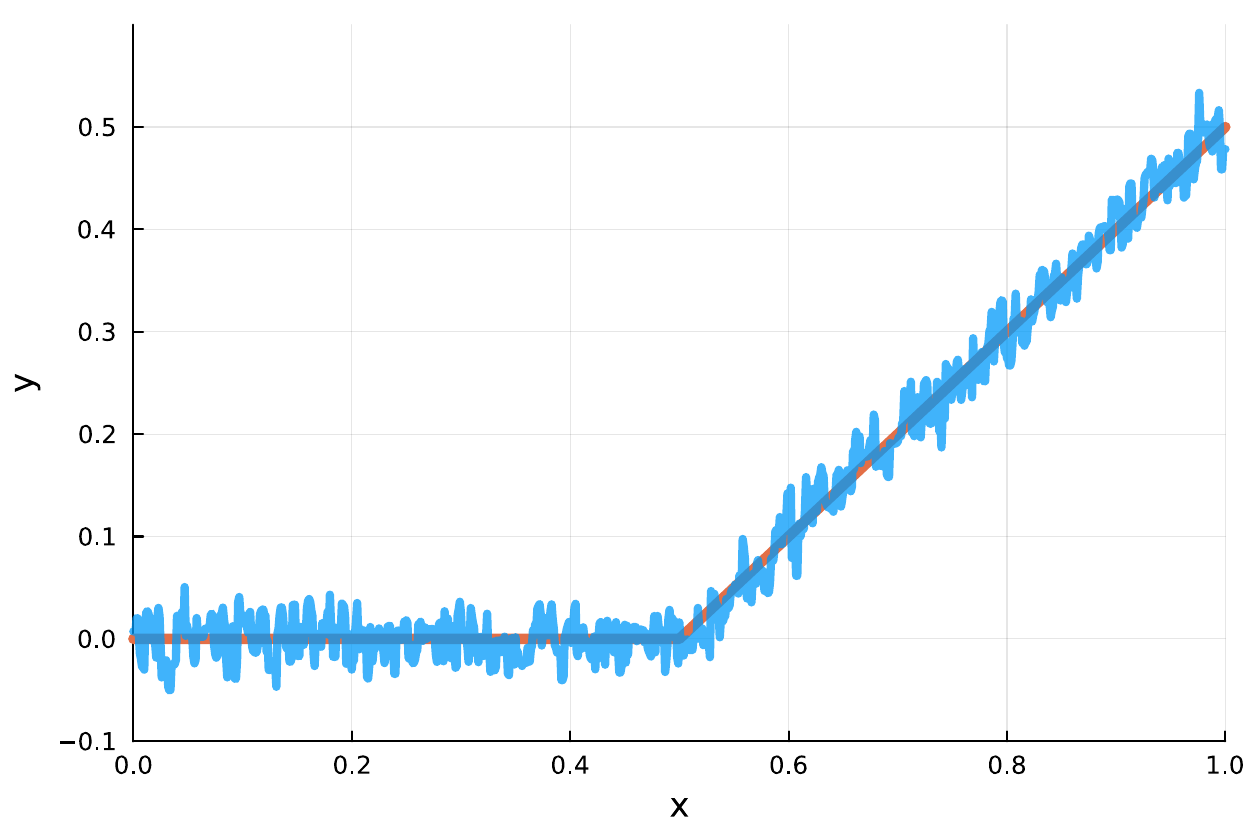}
      \label{fig:SNR_Tree}
    \end{subfigure}
    \begin{subfigure}{.4\textwidth}
      \centering
      \includegraphics[width=\linewidth]{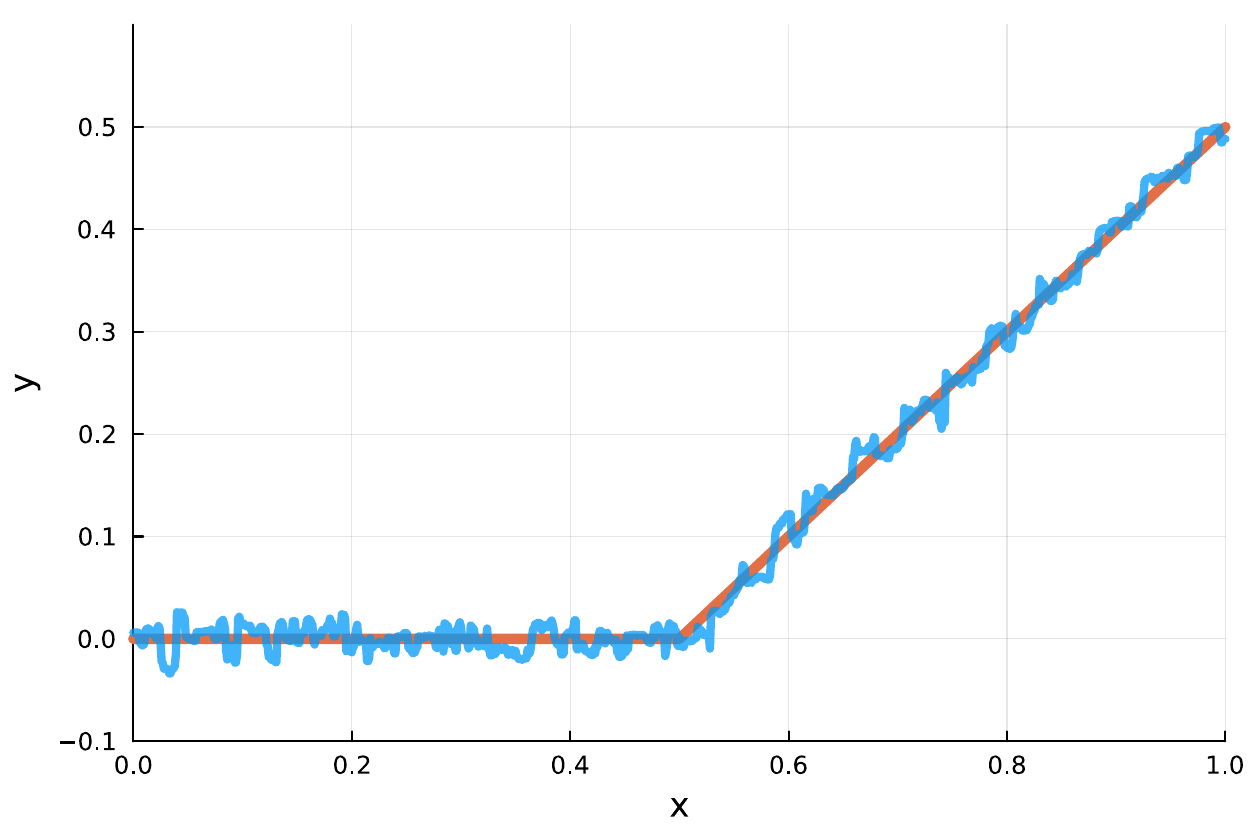}
      \label{fig:SNR_RF25}
    \end{subfigure}
    \begin{subfigure}{.4\textwidth}
      \centering
      \includegraphics[width=\linewidth]{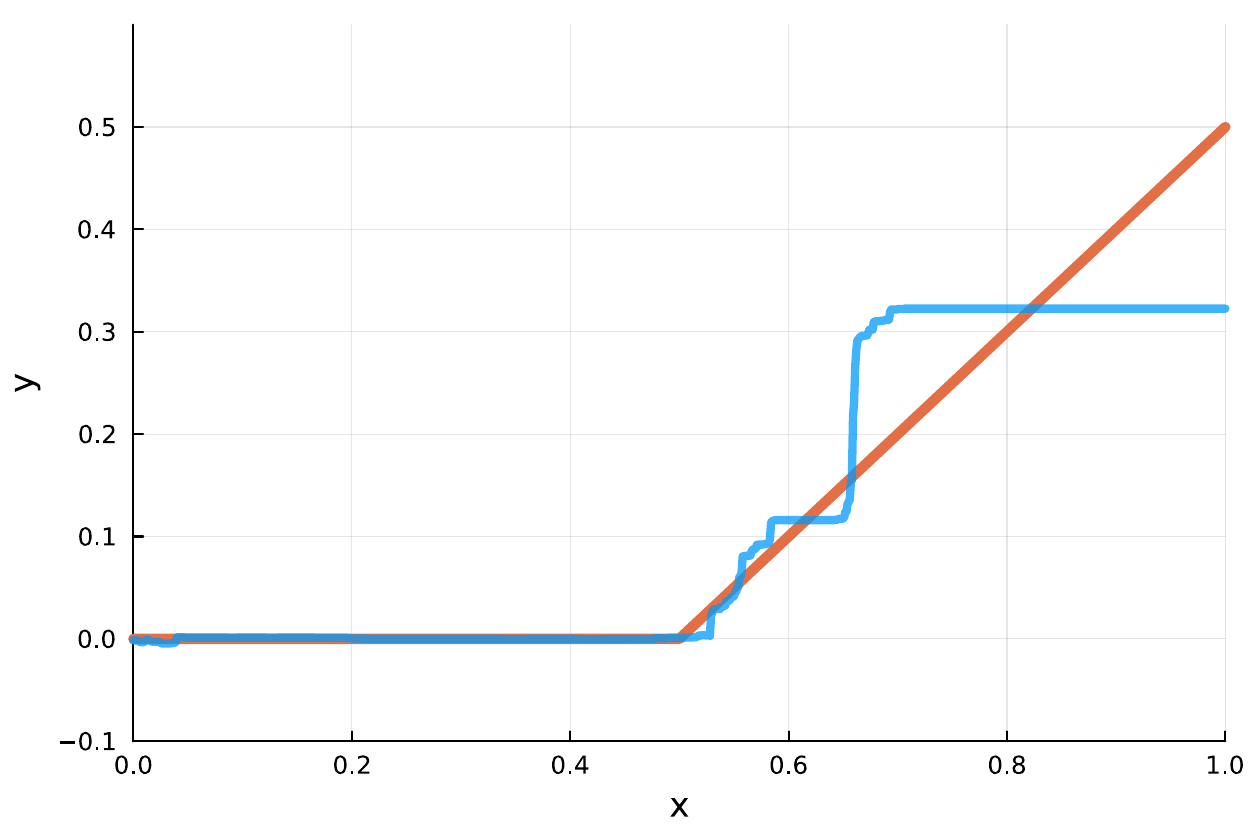}
      \label{fig:SNR_RF500}
    \end{subfigure}
    \begin{subfigure}{.4\textwidth}
        \centering
        \includegraphics[width=\linewidth]{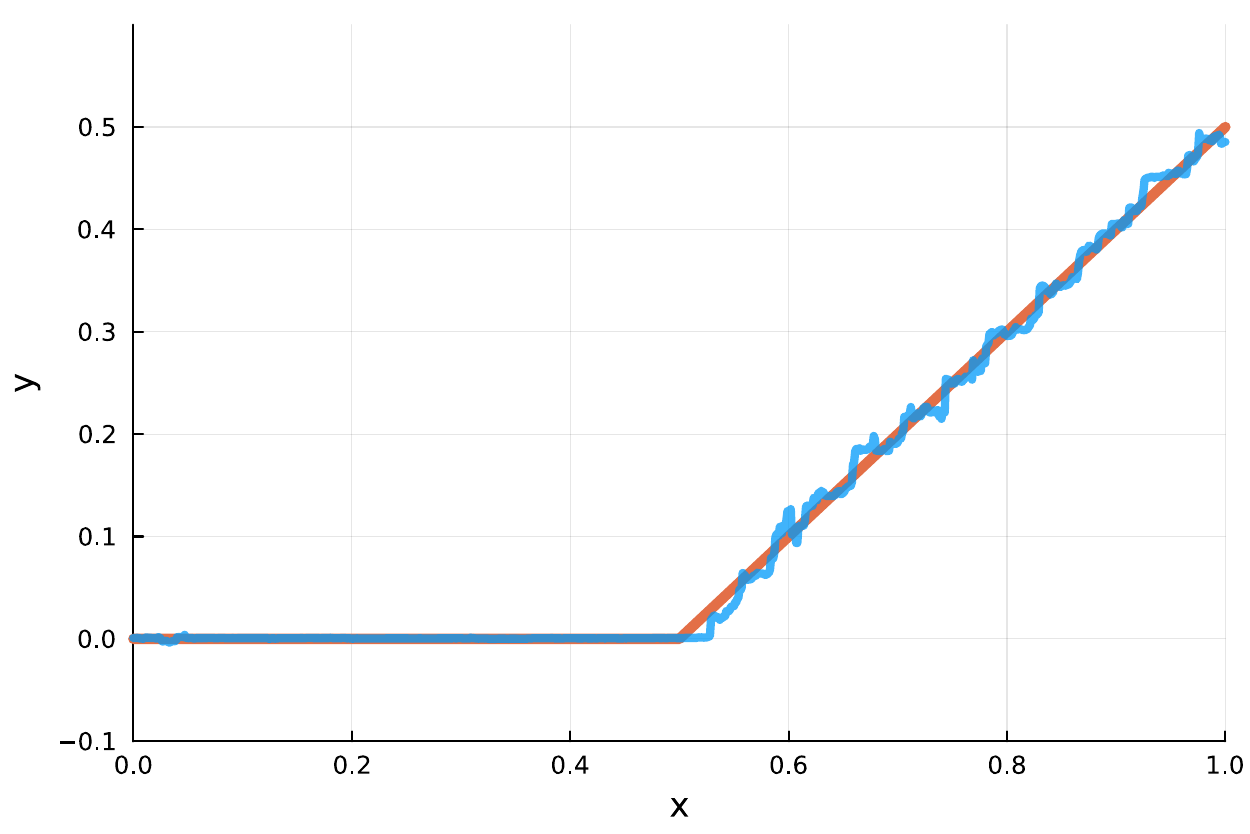}
        \label{fig:SNR_RFtrim}
      \end{subfigure}
    \caption{\textbf{Top left to bottom right:} Predicted values from \texttt{RF-5}, 
    \texttt{RF-25}, \texttt{RF-500}, and \texttt{AlphaTrim} applied 
    to a modified version of the \texttt{elbow} data.}
    \label{fig:SNR_results}
\end{figure}

From our experiments it is apparent that tree sizes within the RF procedure 
should be tuned any time there is concern that the SNR is not very large.  
This is in contrast to some common recommendations in the literature. 
Additionally, alpha-trimmed RFs produce viable fits for the models we consider at 
all SNR levels. \cref{fig:SNR_results} additionally shows why $n_\text{min}$ tuning can be 
inadequate, as such an approach can fail to adapt to surfaces with both flat and 
changing patterns (mixed SNR), whereas alpha-trimmed RFs generally adapt better to such structures.

\subsection{Data sets}

\begin{figure}[t!]
    \centering
    \includegraphics[width=0.8\textwidth]{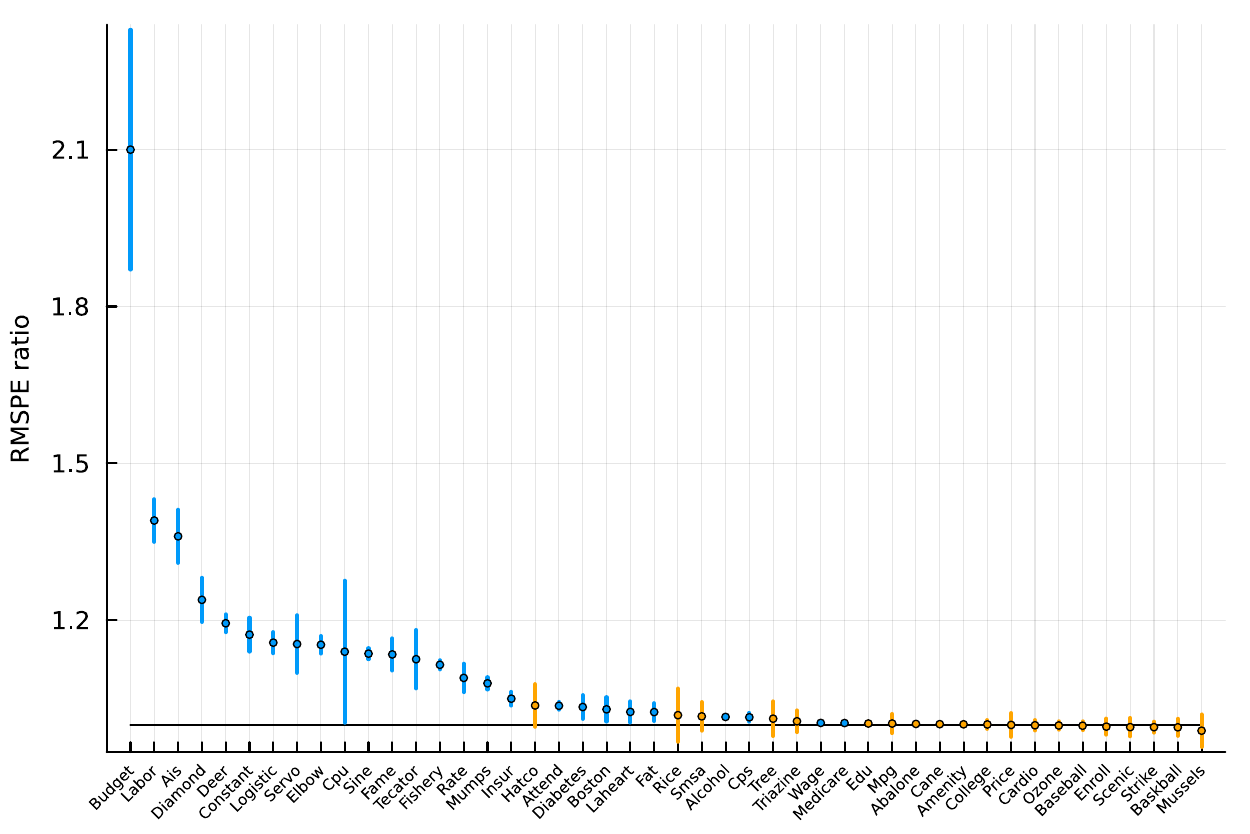}
    \caption{Comparison of alpha-trimmed RFs and default RFs on 46 data sets with approximate 
    95\% $z$-based confidence intervals of ratios of RMSPEs.
    \textcolor{juliablue}{Blue}, \textcolor{juliaorange}{orange}, and 
    \textcolor{juliared}{red} bars indicate cases where the alpha-trimmed RF
    performed better than, similar to, or worse than the default RF.}
    \label{fig:data_sets_RF5_RFtrim}
\end{figure}

\begin{figure}[t!]
    \centering
    \includegraphics[width=0.8\textwidth]{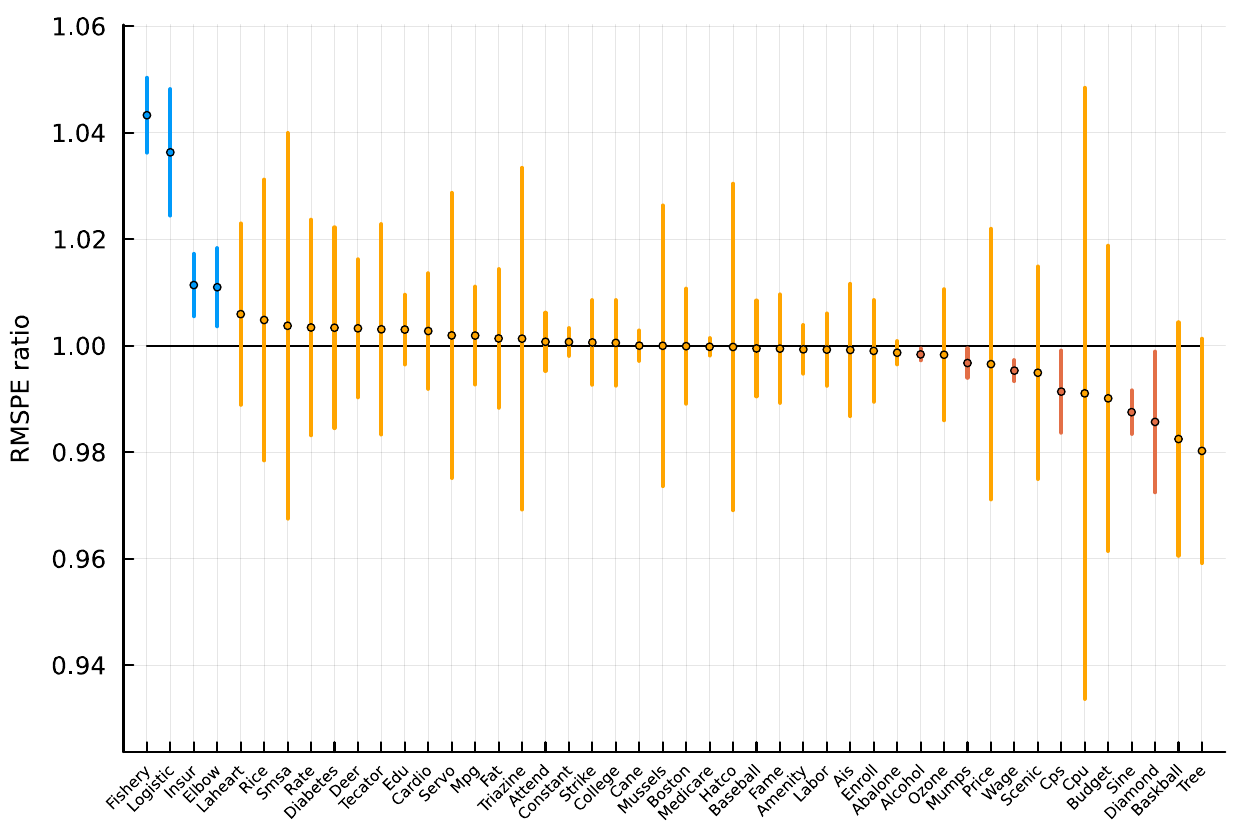}
    \caption{Comparison of alpha-trimmed RFs and tuned RFs on 46 data sets.}
    \label{fig:data_sets_RFBest_RFtrim}
\end{figure}

We examine the performance of our alpha-trimmed RF compared to default and tuned RFs on 
46 data sets.
\cref{fig:data_sets_RF5_RFtrim} shows a comparison of the default RF with 
$n_\text{min} = 5$ and $d'=d/3$. 
The vertical axis is the RMSPE of the default RF divided by the RMSPE of the alpha-trimmed RF. 
The dot represents the average across 10 repetitions of six-fold cross-validation.
Approximate 95\% $z$-based confidence intervals are added 
based on estimates from the 10 replicates. We see that the alpha-trimmed RF is 
frequently favourable compared to the default RF and never significantly worse.

In \cref{fig:data_sets_RFBest_RFtrim}, a comparison is made 
between the alpha-trimmed RF and the standard RF with tuned $n_\text{min}$. 
In this case we see that the alpha-trimmed RF performs comparably to the tuned RF.  
Neither method is significantly better than the other in the majority of the data sets 
in \cref{fig:data_sets_RFBest_RFtrim}.
One possible explanation for this is that the 42 non-synthetic data sets of \cite{chipman2010bart} 
may not have a greatly varying SNR across the domain.
In any case, we again note that even an approach as simple as 
the tuning of $n_\text{min}$ is seldom recommended in practice and 
these simulation results suggest that there can be merit in tuning the depth of trees 
both locally and globally.

\section{Conclusion}
\label{sec:conclusion}
In this work we considered an adaptive approach to pruning individual regression trees
in a random forest. Although pruning is contrary to the conventional 
wisdom that trees should be fully grown, we found that our proposed alpha-trimming
algorithm resulted in a mean squared prediction error
that was often significantly lower, and never significantly higher,
when compared to default random forests with fully-grown trees across 46 example data sets.
Our alpha-trimming algorithm contains a tuning parameter that controls the amount of
pruning to be performed to each tree and the fully-grown random forest is a special case
of this algorithm. The selection of the tuning parameter and the pruning procedure
\textit{does not} require cross-validation and can instead be done
using out-of-bag observations. We therefore advocate our method as an efficient
approach to pruning regression trees in a random forest adaptively, where 
more pruning is applied in regions with a low signal-to-noise ratio.

\acks{We acknowledge the support of the Natural Sciences and Engineering Research 
Council of Canada (NSERC), [funding reference number RGPIN-2018-04868]. 
N.S. acknowledges the support of a Vanier Canada Graduate Scholarship.}

\clearpage

\appendix
\section*{Appendix A: Proofs}
\label{sec:proofs}
In this section we prove the theoretical results presented in \cref{sec:theory}.
Throughout, we use various properties of quadratic forms. (For a resource we refer the reader to 
\cite{searle2016linear}.)

\bprfof{\cref{prop:BIC_consistency_1}}
We proceed by recognizing that a weighted difference of the log-likelihoods
for the tree root and stump models (when a split-point location is fixed) 
is distributed according to a chi-squared random variable. 
To determine the probability that one BIC is greater than another for the 
two models, we show that the probability that a chi-squared random variable 
is greater than the penalty (which is proportional to $\log n$) decays appropriately
so that a union bound over the possible split locations still yields the desired 
consistency result.

Define the quantities
\[ 
  V_{n_L,j} &= \sum_{i=1}^n (Y_i -  \bar{Y})^2 - 
    \sum_{i=1}^{n_L} (Y_i(j) - \bar{Y}_{L,n_L}(j))^2 - 
     \sum_{i=n_L + 1}^n (Y_i(j) - \bar{Y}_{R,n_L}(j))^2, \\ 
  S_{n_L, j}(\sigma^2) &= \left\{3\log(n) \leq \frac{1}{\sigma^2} V_{n_L,j} \right\} \\
  S(\sigma^2) &= \bigcup_{\substack{n_L = 1, \ldots, n-1 \\ j=1,\ldots, d}} S_{n_L,j}(\sigma^2)
    = \cbra{ 3 \log(n) \leq \max_{\substack{n_L = 1, \ldots, n-1 \\ j=1,\ldots, d}} 
      \frac{1}{\sigma^2} V_{n_L,j} }.
\] 
Note that $V_{n_L, j}/\sigma^2_0 \sim \chi^2_1$ where $\sigma^2_0$ is the true variance 
under the tree root model \citep{searle2016linear}.

We begin by assuming that $\sigma^2_0$, the true variance under the tree root model 
is known. Letting $Z \sim \distNorm(0,1)$, we have
\[
  \Pr(S_{n_L,j}(\sigma^2_0)) 
  = \Pr(Z^2 \geq 3 \log(n)) 
  \leq \exp(- (3/2) \log(n)) 
  = n^{-3/2}.
\]
Therefore,
\[
  &\Pr(\text{BIC}_1(Z_1, \ldots, Z_n; \sigma^2_0) > \text{BIC}_0(Z_1, \ldots, Z_n; \sigma^2_0)) \\
  &{\quad} = 1 - \Pr(S(\sigma^2_0)) \\
  &{\quad} \geq 1 - \sum_{\substack{n_L = 1, \ldots, n-1 \\ j=1,\ldots, d}} \Pr(S_{n_L,j}(\sigma^2_0)) \\
  &{\quad} \geq 1 - \frac{(n-1)d}{n^{3/2}}.
\]
For fixed $d$, as $n \to \infty$ we have the desired consistency result with known $\sigma^2_0$. 

We now show that the result still holds when $\sigma^2_0$ is replaced with a
consistent estimator $\hat{\sigma}_n^2 \xrightarrow{p} \sigma^2_0$.
Because $\sigma^2_0 > 0$, we have $1/\hat{\sigma}^2_n \xrightarrow{p} 1/\sigma^2_0$. 
From here, set $\eps > 0$ sufficiently small, for example $\eps = 0.2/\sigma^2_0$. 
We also have
$\Pr(\abs{1/\hat{\sigma}^2_n - 1/\sigma^2_0} \leq \eps) \to 1$ as $n \to \infty$.
Then, define 
\[
  V 
  &= \max_{\substack{n_L = 1, \ldots, n-1 \\ j=1,\ldots, d}} 
    \sbra{ \sum_{i=1}^n (Y_i -  \bar{Y})^2 
      - \sum_{i=1}^{n_L} (Y_i(j) - \bar{Y}_{L,n_L}(j))^2 
      - \sum_{i=n_L + 1}^n (Y_i(j) - \bar{Y}_{R,n_L}(j))^2 }.
\]
Note that $V \geq 0$. Then,
\[
  \MoveEqLeft[3]
  \Pr(\text{BIC}_1(Z_1, \ldots, Z_n; \hat{\sigma}^2_n) > 
    \text{BIC}_0(Z_1, \ldots, Z_n; \hat{\sigma}^2_n)) \\
  &\geq \Pr\rbra{ \cbra{ \frac{1}{\hat{\sigma}^2_n} V < 3 \log(n) } \cap 
    \{ \abs{1/\hat{\sigma}^2_n - 1/\sigma^2_0} \leq \eps \} } \\
  &= \Pr\rbra{ \cbra{ \rbra{ \frac{1}{\sigma^2_0} + \frac{1}{\hat{\sigma}^2_n} - 
    \frac{1}{\sigma^2_0} } V < 3 \log(n) } \cap \{ |1/\hat{\sigma}^2_n - 
      1/\sigma^2_0| \leq \eps \} } \\
  &\geq \Pr\rbra{ \cbra{ \rbra{ \frac{1}{\sigma^2_0} + \eps } V < 3 \log(n) } 
    \cap \{ |1/\hat{\sigma}^2_n - 1/\sigma^2_0| \leq \eps \} } \\
  &= \Pr\rbra{ \left\{\frac{1}{\sigma^2_0} V < 2.5 \log(n)\right\} 
    \cap \{ |1/\hat{\sigma}^2_n - 1/\sigma^2_0| \leq \eps \} } \\
  &\to 1,
\]
as $n \to \infty$ because the probability of each event approaches one. 
(The argument for the first event is the same as in the first part of the proof with 
known $\sigma^2_0$.)
\eprfof

\bprfof{\cref{prop:BIC_consistency_2}}
We first prove the result assuming that $\sigma^2_1$, the true variance 
under the tree stump model, is known. Let $N_L = \sum_{i=1}^n \mathbbm{1}(X_i^{(j)} < l)$
be the number of observations strictly to the left of a fixed $l$ along the 
$j^\text{th}$ variable. Defining $N_R = n-N_L$, set 
\[
  \bar{Y}_{L,l} = N_L^{-1} \sum_{i=1}^n Y_i \mathbbm{1}(X_i^{(j)} < l), \qquad
  \bar{Y}_{R,l} = N_R^{-1} \sum_{i=1}^n Y_i \mathbbm{1}(X_i^{(j)} \geq l).
\] 
Finally, define 
\[
  D 
  = \sum_{i=1}^n (Y_i - \bar{Y})^2 - 
    \sum_{i=1}^n (Y_i - \bar{Y}_{L, l})^2 \mathbbm{1}(X_i^{(j)} < l) - 
    \sum_{i=1}^n (Y_i - \bar{Y}_{R, l})^2 \mathbbm{1}(X_i^{(j)} \geq l).
\] 
From this split, we see that
\[    
  \Pr(\text{BIC}_0(Z_1, \ldots, Z_n; \sigma^2_1) > \text{BIC}_1(Z_1, \ldots, Z_n; \sigma_1^2)) 
  \geq \Pr(\{D/\sigma_1^2 > 3 \log(n)\} \cap \{ N_L \notin \{0, n\} \}).
\]

Under our assumptions, $\Pr(N_L \notin \{0, n\}) \to 1$ as $n \to \infty$.
We proceed to show that $\Pr(D/\sigma_1^2 > 3 \log(n)) \to 1$. 
Set 
\[
  G_n = \cbra{N_L < n^{1/2}} \cup \cbra{N_L > n-n^{1/2}},
\] 
so that 
\[
  \label{eq:two_cases_consistency_2}
  \Pr(D/\sigma_1^2 \leq 3 \log(n)) 
  &= \Pr\rbra{ \{D/\sigma_1^2 \leq 3 \log(n)\} \cap G_n } + 
    \Pr\rbra{ \{D/\sigma_1^2 \leq 3 \log(n)\} \cap G_n^c }.
\]
Under our assumptions, and as a consequence of the weak law of large numbers, 
$\Pr(G_n) \to 0$ as $n \to \infty$. Therefore, the first term on the right side 
of \eqref{eq:two_cases_consistency_2} approaches zero. 
For the second term on the right side of \eqref{eq:two_cases_consistency_2},
observe that for a fixed $n_L \in \cbra{1, \ldots, n-1}$,
\[
  \Pr(D/\sigma_1^2 \leq 3 \log(n) \mid N_L = n_L) 
  = \Pr(Q_{\lambda(n_L)} \leq 3 \log(n)),
\]
where $Q \sim \chi^2_1(\lambda(n_L))$, which is a non-central chi-square 
distribution with one degree of freedom and non-centrality parameter
\[
  \lambda(n_L) = \frac{n_L (n-n_L)}{2n \sigma_1^2} (\mu_1 - \mu_2)^2,
\]
using the parameterization presented in \cite{searle2016linear}. 
Note that for $n_L \in [n^{1/2}, n-n^{1/2}]$, the minimum of $\lambda(n_L)$ is 
\[
  \lambda^\star 
  = \lambda(n^{1/2}) 
  = \lambda(n-n^{1/2}) 
  = \frac{n^{1/2} (n-n^{1/2})}{2n \sigma_1^2} (\mu_1 - \mu_2)^2. 
\]
For a sufficiently small $t > 0$, 
\[
  \MoveEqLeft[3]
  \Pr(\{D/\sigma_1^2 \leq 3 \log(n)\} \cap G^c) \\
  \begin{split}
      ={}&  \Pr(\{D/\sigma_1^2 \leq 3 \log(n)\} \cap \{N_L = \ceil{\sqrt{n}}\}) \\
        & + \ldots + \Pr(\{D/\sigma_1^2 \leq 3 \log(n)\} \cap \{N_L = \floor{n-\sqrt{n}}\}) 
  \end{split} \\
  \leq{}& \Pr\rbra{ Q_{\lambda^\star} \leq 3 \log(n) } \\
  ={}& \Pr(-tQ_{\lambda^\star} \geq -3 t \log(n)) \\
  ={}& \Pr(\exp(-tQ_{\lambda^\star}) \geq n^{-3t}) \\
  \leq{}& \Exp(\exp(-tQ_{\lambda^\star})) \cdot n^{3 t} \\
  ={}& \frac{1}{(1+2t)^{1/2}} 
    \exp\cbra{ -\frac{n^{1/2} (n-n^{1/2})}{2n \sigma_1^2} (\mu_1-\mu_2)^2 [1-(1+2t)^{-1}] 
      + 3 t \log(n) } \\
  ={}& \frac{1}{(1+2t)^{1/2}} \exp\cbra{ -\frac{1}{2 \sigma_1^2} n^{1/2} \rbra{1-\frac{1}{\sqrt{n}}} 
    (\mu_1 -\mu_2)^2 [1-(1+2t)^{-1}] + 3 t \log(n) } \\
  \to{}& 0,
\]
as $n \to \infty$, because $\abs{\mu_1 - \mu_2} > 0$ and $1-1/(1+2t) > 0$.

We now show that the consistency result still holds when $\sigma^2_1$ is replaced with a 
consistent estimate $\hat{\sigma}^2_n$.
Let $D \geq 0$ be as above and note that $\Pr(N_L \notin \{0,n\}) \to 1$. 
As for the proof of \cref{prop:BIC_consistency_1} with $\hat{\sigma}^2_n$, 
set $\eps = 0.2/\sigma^2_1$, for example. We then have 
\[ 
  \MoveEqLeft[3]
  \Pr(\text{BIC}_0(Z_1, \ldots, Z_n; \hat{\sigma}^2_n) > 
    \text{BIC}_1(Z_1, \ldots, Z_n; \hat{\sigma}^2_n)) \\
  &\geq \Pr\rbra{ \cbra{ \frac{1}{\hat{\sigma}^2_n} D > 3 \log(n) } \cap 
    \cbra{N_L \notin \{0, n\}} \cap 
    \cbra{\abs{1/\hat{\sigma}^2_n - 1/\sigma^2_1} \leq \eps} } \\
  &= \Pr\rbra{ \cbra{\rbra{ \frac{1}{\sigma^2_1} + \frac{1}{\hat{\sigma}^2_n} - 
    \frac{1}{\sigma^2_1} } D > 3 \log(n) } \cap 
    \cbra{ N_L \notin \{0, n\} } \cap 
    \cbra{ \abs{1/\hat{\sigma}^2_n - 1/\sigma^2} \leq \eps }} \\
  &\geq \Pr\rbra{ \cbra{\rbra{\frac{1}{\sigma^2_1} - \epsilon} D > 3 \log(n)} \cap 
    \{ N_L \notin \{0, n\} \} \cap 
    \{ \abs{1/\hat{\sigma}^2_n - 1/\sigma^2} \leq \eps \} } \\
  &= \Pr\rbra{ \cbra{\frac{1}{\sigma^2_1} D > 3.75 \log(n) } \cap 
  \{ N_L \notin \{0, n\} \} \cap 
  \{ \abs{1/\hat{\sigma}^2_n - 1/\sigma^2} \leq \eps \} } \\
  &\to 1,
\]
as $n \to \infty$, by the same argument as above for known $\sigma^2_1$. 
\eprfof

\bprfof{\cref{prop:SNR_MSPE}}
We show that for any $j \in \cbra{0,1,\ldots,k-1}$, 
\[
  \EE[(\hat{\mu}(X) - \mu(X))^2] 
  = \frac{\sigma^2}{m} + \rbra{1 + \frac{1}{m}} \cdot \frac{\beta^2}{12 k^2}, 
  \qquad X \sim U([j/k, (j+1)/k]).
\]
Without loss of generality suppose $j = 0$ and consider any $X \in [0, 1/k)$.
Note that 
\[
  \EE[\hat{\mu}(X) \mid X] - \mu(X)
  = \EE[m^{-1} (Y_1 + \cdots + Y_m)] - \beta X
  = \beta \rbra{\frac{1}{2k} - X}.
\]
Similarly, 
\[
  \Var[\hat{\mu}(X) \mid X]
  = \frac{1}{m} \rbra{\sigma^2 + \frac{\beta^2}{12 k^2}}.  
\]
Therefore,
\[
  \EE[(\hat{\mu}(X) - \mu(X))^2 \mid X]
  &= (\EE[\hat{\mu}(X) \mid X] - \mu(X))^2 + \Var[\hat{\mu}(X) \mid X] \\
  &= \left( \beta \rbra{\frac{1}{2k} - X} \right)^2 + 
    \frac{1}{m} \rbra{\sigma^2 + \frac{\beta^2}{12 k^2}}.
\]
Taking expectations over the randomness in $X \sim U([0, 1/k))$ yields the 
desired result.
\eprfof

\bprfof{\cref{prop:comp_complexity}} 
The results are based on work in \cite{henrey2016statistical}.
\eprfof
\section*{Appendix B: Additional experimental details}

\subsection*{Generation of synthetic data} 
Here we describe the generation of the four synthetic data sets used in our simulation study: 
\texttt{constant}, \texttt{elbow}, \texttt{logistic}, and \texttt{sine}.

In all settings we set $n = 1,000$ and $X_1, \ldots, X_n \stackrel{iid}{\sim} U(0, 1)$, 
with $Y_i | X_i \sim \distNorm(\mu(X_i), \sigma^2)$ for some values of $\mu$ and 
$\sigma^2$.
For the \texttt{constant} dataset, we have $\mu(x) = 0$ and $\sigma^2 = 1/1000$.
For the \texttt{elbow} dataset, we have $\mu(x) = 0$ if $x < 0.5$ and $\mu(x) = x-0.5$ if 
$x \geq 0.5$, with $\sigma^2 = 1/1000$. 
For the \texttt{logistic} dataset, we use $\mu(x) = 1/(1+\exp(15-30x))$ and $\sigma^2 = 0.005$. 
Finally, for the \texttt{sine} dataset, we have $\mu(x) = \sin(2\pi x)$ and $\sigma^2 = 0.05$. 
After generating the data, the responses $Y_i$ are normalized to have mean zero 
and unit variance.

\subsection*{Additional alpha-trimming implementation details}
On occasion, estimates of $\hat\sigma^2$ for a sub-tree are very small 
(e.g., when the tree follows the training data too closely). 
When $\hat\sigma^2 < 10^{-15}$, we replace the variance estimate with 
$\hat\sigma^2_\text{terminal}/2$, where $\hat\sigma^2_\text{terminal}$ is the estimate 
of the variance under the assumption that the given node is a terminal node. 
If $\hat\sigma^2$ is still below the threshold of $10^{-15}$, we throw an error 
and suggest to increase the \texttt{min.node.size} parameter.

\vskip 0.2in
\bibliography{main.bib}


\end{document}